\documentclass[runningheads]{llncs}

\usepackage[T1]{fontenc}
\usepackage[british]{babel}
\usepackage{graphicx} 
\usepackage{csquotes}
\usepackage[pdftex,bookmarks,colorlinks]{hyperref}
\usepackage{amsmath}
\usepackage{amssymb}  
\usepackage{subcaption} 
\usepackage[
    style=numeric,
    backend=biber,
    sorting=none,
    doi=true,
    isbn=false,
    url=true
    ]{biblatex}

\begin{document}

    \title{On the Benefits of Uncertainty Estimation for Species Classification: A Case Study on Optimizing Expert re-Annotation}
    \titlerunning{On the Benefits of Uncertainty Estimation for Species Classification}
    
    \author{Kamenan N'gadi\inst{2} \and
    Martin Laviale \inst{1} \and 
    Cédric Pradalier\inst{3} \and 
    Jérémy Fix \inst{2}} 
    
    \institute{
    Université de Lorraine, CNRS, LIEC, F-57000 Metz, France \and
    Université de Lorraine, CentraleSupélec, CNRS, Loria, F-57000 Metz, France \and
    GeorgiaTech Europe, CNRS IRL 2958, F-57000, Metz, France}

    \maketitle 
    
    \begin{abstract}
Automated identification of diatoms from microscopic images is a cornerstone of large-scale freshwater biomonitoring, yet the high diversity of species and strong class imbalance pose significant challenges for reliable classification. In this study, we develop a ResNet-18 baseline for a 759-class diatom dataset and investigate how calibrated confidence estimates support selective classification to reduce expert workload without sacrificing taxonomic integrity. While the initial model exhibits overconfidence with an Expected Calibration Error (ECE) of 0.117, post-hoc temperature scaling ($T=2.12$) dramatically improves reliability, lowering the ECE to 0.055.By leveraging these calibrated probabilities within a triage framework, we demonstrate that a Selective Macro-F1 score of 99.01\% can be maintained at 66.2\% coverage (rejection threshold $\tau = 0.934$). Qualitative analysis reveals that the system effectively functions as an intelligent filter, consistently deferring rare taxa—such as AAMB and AASI—and ambiguous morphologies for expert review. Our findings, supported by the torch-uncertainty framework, highlight the potential of simple calibration techniques to make automated diatom identification practical and robust, prioritizing the preservation of rare species metrics essential for ecological assessment.
To support reproducibility, our code, environment specifications, and calibration results are available anonymously at \href{https://anonymous.4open.science/r/greenprwksp-B53D/README.md}{\url{https://anonymous.4open.science/r/greenprwksp-B53D/README.md}}
\end{abstract}
    \begin{section} {Introduction} 
    
            The over-exploitation of aquatic resources has made robust bio-monitoring tools paramount. Diatoms ---micro-algae are ideal bioindicators due to their ubiquity and sensitivity to environmental variables like pH and nutrient levels, as recognized by the EU Water Framework Directive \cite{du_buf_automatic_2002}. Standard bio-monitoring protocols rely fundamentally on  taxonomic identification at the species level, traditionally performed via microscopy. 
            However, traditional monitoring is hampered by two bottlenecks: a \textbf{taxonomic bottleneck} caused by \textbf{high morphological variability}, and a \textbf{human resource bottleneck} due to the global \textbf{decline in specialized experts}. Consequently, the burden of routine environmental monitoring falls on a shrinking pool of experts, leading to long turnaround times for samples and high costs for regulatory agencies. \cite{du_buf_automatic_2002}\\
            
            However their identification presents significant challenges for automation. First, the MacDonald-Pfitzer rule \cite{mann_species_1999} dictates that cell size and shape change across a species' life cycle, meaning a classifier must map a "trajectory of shapes" rather than a static template. Second, environmental stressors can induce teratologies (structural malformations) which act as out-of-distribution (OOD) samples. It worth to note that for an identification perspective and in automated pipelines, these OOD samples deviate significantly from the "normal" specimens distribution used during model training.  
            Finally, "cryptic diversity" means morphologically identical specimens may belong to different ecological lineages, introducing inherent label noise and inter-expert disagreement \cite{mann_species_1999}.\\
            Given the inherent subjectivity in diatom taxonomy, identification is prone to significant inter-expert variability; different specialists may reach conflicting conclusions for the same specimen, especially with degraded samples. If the training labels for an automated system are inconsistent, the model’s performance is fundamentally capped. Consequently, a robust monitoring framework must account for this label noise and provide mechanisms to flag cases where expert disagreement is likely.
             
            The effort to \textbf{automate diatom identification} has transitioned from early computer vision techniques \cite{du_buf_automatic_2002} to modern deep learning architectures. 
            While recent Deep Learning (DL) models approach human-level accuracy on curated datasets \cite{kloster_deep_2020}, they struggle with real-world noise and tend to be "overconfident." Standard Softmax probabilities often mask underlying uncertainty, which is unacceptable in high-stakes ecological diagnostics \cite{kendall_what_2017}.   
             
             To address this, we need systems that not only classify but also \textbf{estimate the reliability of their predictive uncertainty}. Selective classification (learning with rejection) \cite{geifman_selective_2017} provides a formal paradigm: a confidence threshold decides whether a sample is automatically processed or deferred to a human. By trading off coverage (fraction of samples handled automatically) against selective risk (error rate on those samples), such systems can guarantee that expert involvement is reserved for the most ambiguous cases.
             Effective selective classification requires a nuanced understanding of predictive uncertainty. Uncertainty \cite{hullermeier_aleatoric_2021} is not a monolithic concept; rather, it arises from different sources that have different implications for the monitoring process.
             \textbf{Aleatoric uncertainty} ---irreducible noise from imaging artifacts, morphological overlap or label ambiguity.
            Because aleatoric uncertainty is a property of the data, it cannot be reduced by simply collecting more training samples of the same type. Instead, the model must be able to recognize when an image is fundamentally ambiguous and signal this "irreducible noise" to the system.
            In the other hand, \textbf{Epistemic uncertainty}---model ignorance in data-sparse regions of the data-generating process. In diatom monitoring, epistemic uncertainty is typically triggered by rare species or new imaging conditions.\\
            
            
            In this study, we adopt the framework of selective classification for diatom re-annotation by human expert. We equip a deep classifier with a rejection mechanism based on temperature-scaled softmax outputs \cite{guo_calibration_2017}. By defining a confidence threshold via risk-coverage curves, we ensure that automated identifications stay below a target error rate (e.g., 1\%), deferring ambiguous cases to human experts. 
            We view this as a first step towards fully uncertainty‑aware diatom monitoring. 
            This lightweight approach provides a practical bridge between high-throughput imaging and reliable ecological assessment and ultimately can meaningfully reduce expert workload.

    \end{section}

    \section{Related Works}
    Our work lies at the intersection of four research domains: Automated diatom classification, uncertainty quantification (UQ) in deep learning, selective classification, conformal prediction for human-in-the-loop systems, and active learning.\\

    The automation of diatom identification has been pursued for several decades.
    Early systems such as ADIAC~\cite{du_buf_automatic_2002} demonstrated that automated identification even for challenging species was feasible, achieving accuracies between 75\% and 90\% using classical computer vision techniques (e.g., hand-crafted contour and texture features).
    Since then, deep learning has become the dominant paradigm.
    Recent studies have achieved impressive classification performance.
   Kloster~\textit{et~al.}~\cite{kloster_deep_2020} showed that a VGG16-based CNN reached F1~scores above 97\% for species-level classification on virtual slides, demonstrating that domain adaptation to diatoms is viable with limited annotated data.
    Building on such advances, Venkataramanan~\textit{et~al.}~\cite{venkataramanan_gaussian_2023} proposed a method that enforces a Gaussian structure in the latent space of a deep classifier; approximate Gaussian is obtained through a self‑supervised procedure that dynamically splits non‑Gaussian classes into multiple Gaussian sub‑classes.
    With a well‑regularised latent space, the Mahalanobis distance to the corresponding class centroid serves as a principled uncertainty score, enabling both out‑of‑distribution sample detection and improved probability calibration.
    Their approach requires minimal architectural modifications and was evaluated on diatoms, making it directly relevant to diatom identification.
    A complementary direction addresses the scarcity of annotated data: \cite{tan_scaling_2025} introduced self-supervised pre-training for diatom classification, showing that fine-tuning with only 50 samples per class can recover full-supervised accuracy, reducing annotation requirements by roughly 96\%.
    These works demonstrate that deep learning can match or exceed human-level accuracy on curated benchmarks.
    However, they focus primarily on maximising classification performance rather than on the reliability of predictions in operational settings.
    None of these systems provides a principled mechanism for deciding when a prediction should be trusted versus deferred to an expert.  \\ 
    
    As discussed in the introduction, the machine learning community distinguishes two sources of uncertainty in neural network predictions: aleatoric (irreducible data noise) and epistemic (model ignorance that can be reduced with more data or better modelling)~\cite{kendall_what_2017,hullermeier_aleatoric_2021}.
    This distinction is critical for bioindication—a model trained on typical diatom specimens may exhibit low epistemic uncertainty for common species but high epistemic uncertainty when faced with teratological forms, which are precisely the cases that warrant expert review.\\
    
    Several techniques have been developed to estimate epistemic uncertainty. Monte Carlo (MC) Dropout~\cite{gal_dropout_2016} treats dropout at test time as a Bayesian approximation: by performing multiple stochastic forward passes and computing the variance of the predictions, one obtains a measure of model uncertainty without any architectural change.
    Deep Ensembles~\cite{lakshminarayanan_simple_2017} take an alternative route: several models are trained independently with different random initializations, and their predictions are aggregated; the disagreement among ensemble members provides a reliable epistemic uncertainty estimate.
    Both approaches, as well as their variants, have been thoroughly surveyed by Gawlikowski~\textit{et~al.}~\cite{gawlikowski_survey_2023}, who cover Bayesian neural networks, ensembles, test-time augmentation, and related calibration aspects.
    Having methods to quantify uncertainty does not directly prescribe how to use them in decision-making.\\
    
    The framework of selective classification (learning with rejection)~\cite{geifman_selective_2017} formalises the trade-off between automated throughput and accuracy through risk--coverage curves that characterise model performance as a function of the fraction of examples on which it abstains.
    Recent extensions include hierarchical selective classification that leverages taxonomic structure to make predictions at coarser ranks when fine-grained confidence is insufficient~\cite{goren_hierarchical_2024}, and extensive benchmarking studies that highlight the dependency of the best method on user objectives~\cite{pugnana_deep_2024}. Beyond binary rejection, a more nuanced triage strategy involves providing a list of plausible candidates when a single species cannot be uniquely identified.  
    Conformal prediction~\cite{shafer_tutorial_nodate} is a distribution‑free framework that turns any predictor into a set-valued  predictor with finite‑sample coverage guarantees.
    Given a pre‑trained classifier, a calibration set, and a user‑specified error rate $\alpha$, conformal prediction constructs for each test point $x$ a prediction set $\mathcal{C}(x)\subseteq\mathcal{Y}$ such that
    \begin{equation}
    \Pr\bigl[Y_{\text{test}}\in\mathcal{C}(X_{\text{test}})\bigr]\ge 1-\alpha
    \label{eq:conformal_guarantee}
    \end{equation}
    holds in the marginal sense over the random draw of the calibration and test samples.
    The guarantee~\eqref{eq:conformal_guarantee} therefore holds regardless of the underlying data distribution, provided the calibration and test data are exchangeable \cite{shafer_tutorial_nodate}.
    EPICSCORE~\cite{cabezas_epistemic_2025} enhances this framework by injecting an explicit estimate of epistemic uncertainty (obtained via MC Dropout or Gaussian Processes) into the nonconformity score, so that prediction sets automatically expand in data‑poor regions where the classifier is less confident.
    This is particularly well-suited to diatom monitoring, where rare species and teratological forms naturally correspond to data-sparse regions.
    Recent work explicitly connects conformal prediction to human-in-the-loop systems: Horton~\textit{et~al.}~\cite{horton_conformal_2025} proposed Conformal Validation, a deferral policy that uses conformal prediction to identify the most uncertain observations and provides human experts with informative prediction sets.
    
    Our problem also relates to active learning, where the goal is to query a human oracle for labels on the most informative samples in order to improve the model~\cite{li_survey_2025}.
    Active learning aims at building a better classifier over time, whereas our work aims at intelligent \emph{deferral} of decisions in a fixed model, without necessarily retraining.
    While both involve human experts, the objective differs: active learning seeks better models; we seek better workflows that reduce the expert burden in routine monitoring.
    
    To the best of our knowledge, the combination of selective classification with a  calibrated confidence threshold has not yet been systematically explored for diatom identification in operational monitoring scenarios, which motivates the present study.

    \section{Methodology}
    \subsection{Selective classification with a calibrated threshold}

        Routine diatom‑based bioindication requires processing thousands of specimens, yet taxonomic expertise is scarce and time‑consuming. This study therefore seeks to minimise expert involvement by automatically classifying those specimens for which the model is confident, while deferring uncertain cases to a human specialist. Formally, this is cast as selective classification (learning with rejection)~\cite{geifman_selective_2017}: a confidence threshold decides whether a specimen is handled automatically or passed to a taxonomist.
        Let $\mathcal{X}$ be the input space (e.g., diatom images) and $\mathcal{Y} = \{1,\dots,K\}$ the label set. A training sample $\mathbf{S}_m = \{(x_i,y_i)\}_{i=1}^{m} \subseteq \mathcal{X} \times \mathcal{Y}$ is drawn i.i.d.\ from an unknown distribution $\mathbf{P}$ over $\mathcal{X} \times \mathcal{Y}$. A classifier $f: \mathcal{X} \to \mathcal{Y}$ has real risk~:
        \begin{equation}
            R(f) \triangleq \mathbf{E}_{(X,Y)\sim\mathbf{P}}\bigl[\ell(f(X),Y)\bigr],
            \label{eq:true_risk}
        \end{equation}
        where $\ell$ is a loss function (e.g., 0--1 loss).
        A selective classifier is a pair $(f,g)$, with $g:\mathcal{X}\to\{0,1\}$ a selection function that qualifies $f$:
        \begin{equation}
            (f,g)(x) \triangleq
            \begin{cases}
            f(x), & \text{if } g(x)=1,\\[4pt]
            \text{``abstain''}, & \text{if } g(x)=0.
            \end{cases}
            \label{eq:selective_classifier}
        \end{equation}
        Thus the model abstains iff $g(x)=0$.  
        Performance is assessed through coverage and selective risk:
        \begin{align}
            \phi(f,g) &\triangleq \mathbf{E}_{\mathbf{P}}[g(X)], \label{eq:coverage} \\
            R(f,g) &\triangleq \frac{\mathbf{E}_{\mathbf{P}}[\ell(f(X),Y)\,g(X)]}{\phi(f,g)}. \label{eq:selective_risk}
        \end{align}
        From an ecological standpoint, coverage is the fraction of samples processed without human intervention, while selective risk is the expected proportion of misidentified specimens among them—an error that can propagate through multimetric indices and compromise the reliability of an ecological assessment.    
        Given a confidence parameter $\delta>0$ and a target risk $r^*>0$, the aim is to build $g$ from $\mathbf{S}_m$ such that 
        \begin{equation}
            \Pr_{\mathbf{S}_m}\!\bigl(R(f,g) > r^*\bigr) < \delta. \label{eq:sgr_goal}
        \end{equation}
        A standard construction~\cite{geifman_selective_2017} employs a confidence‑rate function $\kappa_f:\mathcal{X}\to\mathbb{R}^+$ that ranks predictions: $\kappa_f(x_1)\ge\kappa_f(x_2)$ means the confidence in $f(x_1)$ is at least that in $f(x_2)$.   As suggest in \cite{geifman_selective_2017} $\kappa_f$. Here we take the maximum softmax probability as in \cite{geifman_selective_2017}.
        The selection function is then parametrised by a threshold $\theta>0$:
        \begin{equation}
            g_\theta(x) \triangleq
            \begin{cases}
                1, & \text{if } \kappa_f(x) \ge \theta,\\
                0, & \text{otherwise.}
            \end{cases}
            \label{eq:selection_function}
        \end{equation}
        Its empirical selective risk on the sample $\mathbf{S}_m$ is
        \begin{equation}
            \hat{r}(f,g \mid \mathbf{S}_m) \triangleq
            \frac{\frac{1}{m}\sum_{i=1}^{m} \ell(f(x_i),y_i)\,g(x_i)}
                 {\hat{\phi}(f,g \mid \mathbf{S}_m)},
            \qquad
            \hat{\phi}(f,g \mid \mathbf{S}_m) \triangleq \frac{1}{m}\sum_{i=1}^{m} g(x_i).
            \label{eq:empirical_selective_risk}
             \end{equation}
        The optimal $\theta$ can be found via the Selection with Guaranteed Risk (SGR) algorithm , which binary‑searches for the threshold that respects the risk guarantee~\cite{geifman_selective_2017}.  
        For the sake of simplicity , in this paper, we consider a very simple approach, compared to the full SGR machinery, to compute the threshold.

        For the curve to yield a trustworthy threshold, the confidence scores must be well‑calibrated: a score $p$ should imply correctness with probability $p$. Modern neural networks, however, are notoriously overconfident~\cite{guo_calibration_2017}, which in the context of biomonitoring would silently increase the fraction of misidentified specimens in the automated set. We therefore apply temperature scaling, a single‑parameter extension of Platt scaling~\cite{guo_calibration_2017}. Given a logit vector $\mathbf{z}$, the calibrated confidence for the predicted class is
        \begin{equation}
            \hat{q} = \max_k\; \sigma_{\mathrm{SM}}\!\left(\frac{\mathbf{z}}{T}\right)^{(k)},
            \label{eq:temp_scaling}
        \end{equation}
        where $\sigma_{\mathrm{SM}}$ is the softmax. The scalar $T>0$ is learnt on the calibration set by minimising the negative log‑likelihood loss. Because rescaling preserves the ranking of the logits, the model’s accuracy is unchanged; only the confidence values are adjusted, enabling a more reliable rejection threshold. A well‑calibrated model thus provides the ecologist with a trustworthy estimate of when it is safe to rely on automated identifications.
        The resulting decision rule directly addresses the taxonomic bottleneck in routine bio-monitoring by reserving expert effort for the most challenging specimens while controlling the error rate on the rest.


            
            

                    \begin{figure}[htbp]
                    \centering
                        \begin{subfigure}[b]{0.48\textwidth}
                            \centering
                            \includegraphics[width=\linewidth]{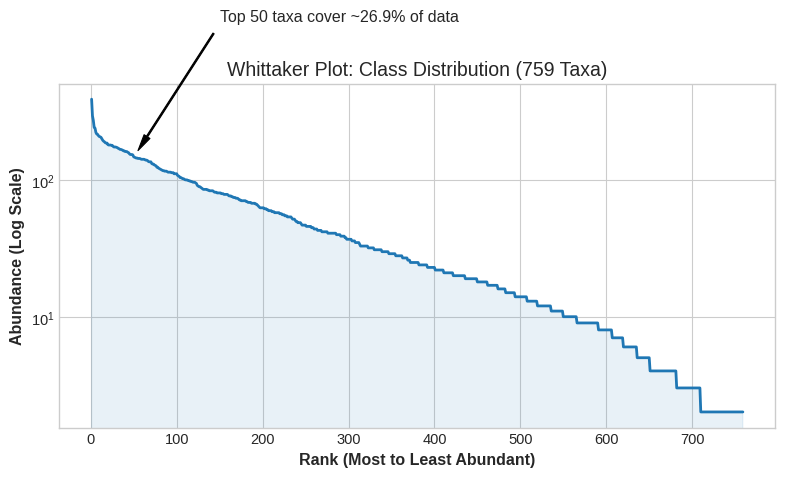}
                            \caption{Taxonomic distribution (Whittaker plot)}
                            \label{subfig:whittaker}
                        \end{subfigure}
                        \hfill
                        \begin{subfigure}[b]{0.48\textwidth}
                            \centering
                            \includegraphics[width=\linewidth]{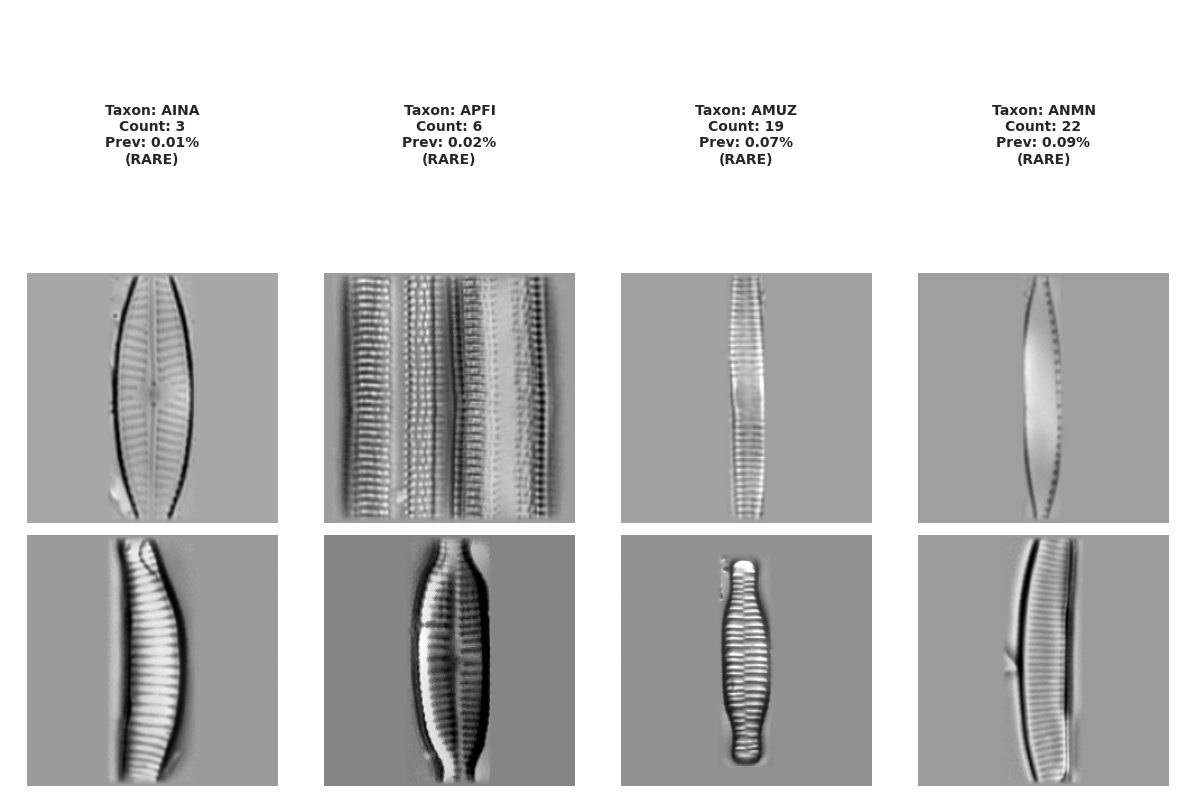}
                            \caption{Morphological variability examples}
                            \label{subfig:samples}
                        \end{subfigure}
                    
                        \caption{\textbf{Characterisation of the DIATLAS dataset: taxonomic imbalance and morphological diversity.} 
                        \textbf{(a)} The plot illustrates the distribution of 35,447 images across 759 taxa on a logarithmic scale. The initial steep decline represents a small group of dominant species, while the extensive "long tail" contains hundreds of rare taxa with sparse representation (fewer than 10 specimens per class). Notably, the top 50 most abundant taxa account for only 26.9\% of the total specimens, indicating that the majority of the dataset consists of low-prevalence classes. 
                        \textbf{(b)} Representative specimens  illustrates a rare taxon ($Prevalence < 0.1\%$) where limited specimen counts—ranging from 3 for AINA to 22 for ANMN—pose a significant challenge for supervised learning. Furthermore, the vertical samples demonstrate high morphological variability, capturing distinct valve and girdle orientations within the same class (e.g., AINA, AMUZ) and variations in frustule preservation and striae visibility (APFI, ANMN).
                        Together, the extreme data scarcity for rare taxa (e.g., CAGL, CIRI) and high visual ambiguity justify the implementation of a selective classification framework to ensure reliable automated bioindication.}
                    \label{fig:dataset_overview}
                \end{figure}

                \subsection{Dataset: DIATLAS}

                    This study uses DIATLAS \cite{galinier_diatlas_2024}, an open dataset of optical microscopy images of diatoms extracted from French regional atlases and documenting freshwater specimens. Images were acquired using optic or photonic microscopy. Each image is associated with a taxonomic classification following the OMNIDIA nomenclature~\cite{lecointe_omnidia_2024}.                   
                    The raw corpus comprises 38,719 images. To ensure statistical viability for deep learning, we retained only classes with at least two specimens, yielding a curated set of 35,447 images across 759 taxa. The resulting class distribution exhibits a pronounced \textbf{long‑tail} shape—an inherent feature of environmental sampling where a few dominant taxa coexist with many rare species—posing a significant challenge for classification (see Figure~\ref{fig:dataset_overview}).
                    
                    For experimentation, the dataset was split into training (25,522 images), calibration (2,835), and test (7,090) subsets via stratified sampling ($72/8/20$). The calibration set is used  as validation set to trigger checkpointing and for temperature scaling and threshold selection. Dataset splits were stratified to preserve class proportions for taxa with sufficient representation; however, for extremely rare species (e.g., those with only one or two instances), this approach naturally resulted in some taxa being present in the test set while remaining absent from the training set or vice-versa.

        \subsection{Experimental Implementation}
            Our framework is built on \verb:torch-uncertainty: (TU) \cite{lafage_torch-uncertainty_2025}. To integrate DIATLAS, we implemented a custom \texttt{DiatomDataModule} with a preprocessing pipeline designed to mitigate covariate shift. Images are rescaled to a target physical resolution, vertically aligned via principal axis rotation, and centre-cropped or padded using seamless cloning with a median-colour background to prevent edge artefacts. Grayscale conversion is applied to isolate structural frustule morphology from acquisition-related colour variance.

            We trained a ResNet-18 classifier for $60$ epochs using Adam ($\text{lr}=10^{-2}$) with a batch size of $32$ and $16$-bit mixed precision. Training utilised gradient accumulation (factor of $4$) and norm-clipping (threshold $1.0$). 
            
            Post-hoc calibration was performed via temperature scaling~\cite{guo_calibration_2017} as implemented in \verb:torch-uncertainty:. While the architecture is standard, the selective classification mechanism was explicitly implemented to provide direct control over the abstention threshold—enabling precise evaluation of coverage--risk trade-offs and class-wise rejection. Performance is quantified via classification accuracy and Expected Calibration Error (ECE) using the maximum softmax probability (MSP). 
            
            All experiments were executed on an NVIDIA RTX 2000 Ada GPU.

    \section{Results}
     \subsection{Calibration and Selective Triage}

\begin{figure}[htbp]
     \centering
     \begin{subfigure}[t]{0.58\textwidth}
         \centering
         \includegraphics[width=\textwidth]{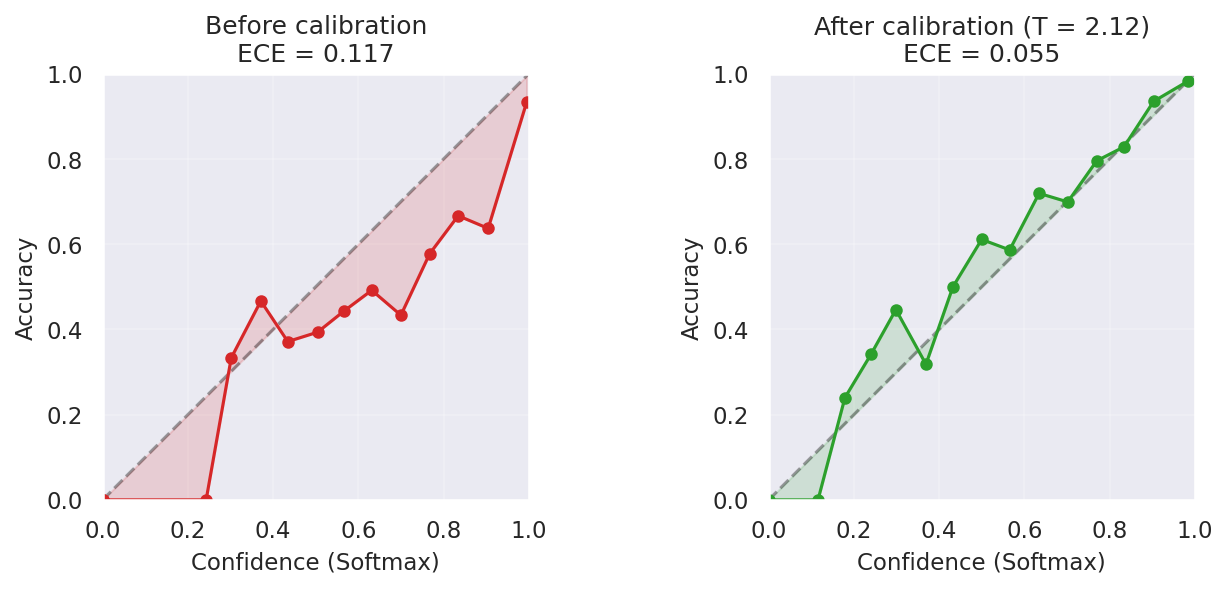}
         \caption{Reliability diagrams (Macro-averaged)}
         \label{fig:reliability_calibration}
     \end{subfigure}
     \hfill
     \begin{subfigure}[t]{0.40\textwidth}
         \centering
         \includegraphics[width=\textwidth]{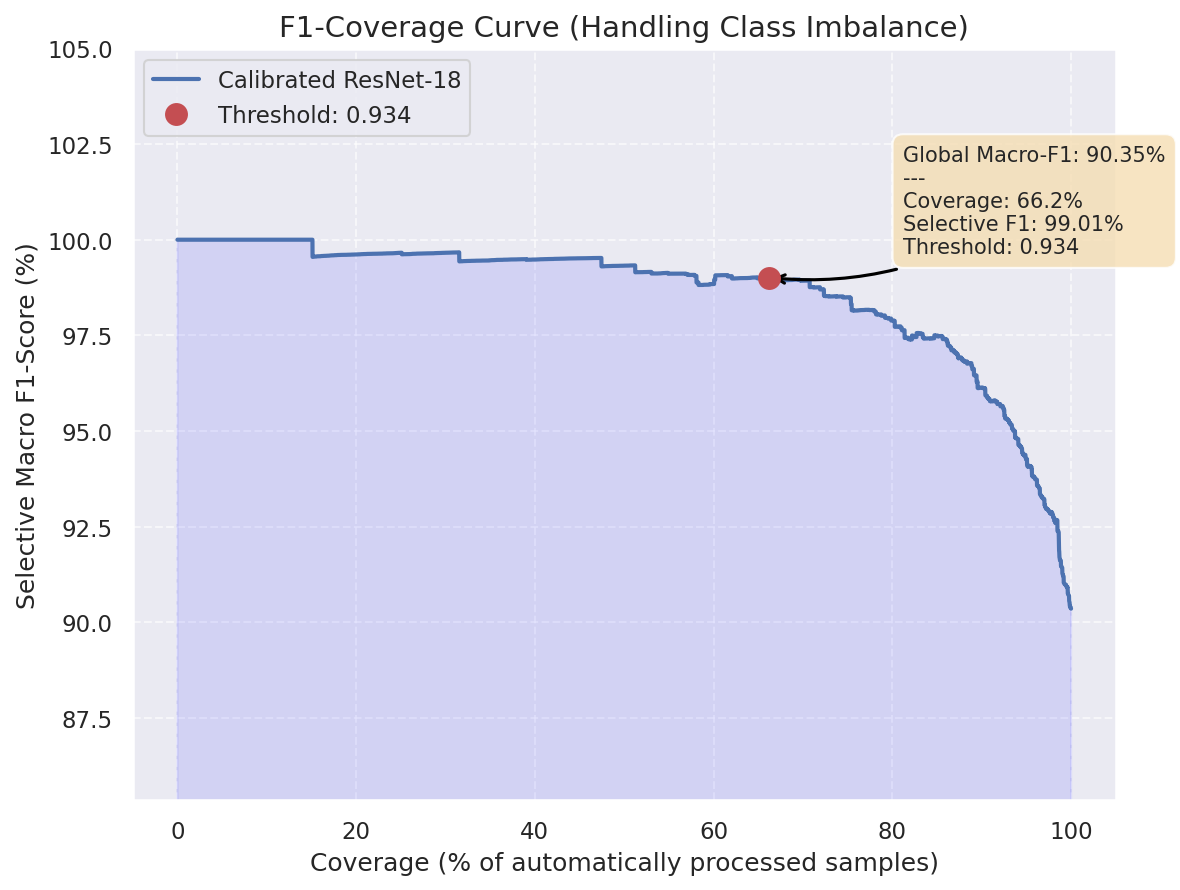}
         \caption{F1-coverage curve}
         \label{fig:risk_coverage}  
     \end{subfigure}
    \caption{\textbf{Calibration and Selective Triage Performance.} \textbf{(a)} Temperature scaling with $T=2.12$ reduces the ECE to $0.055$, correcting overconfidence. \textbf{(b)} The F1-coverage trade-off shows that at a threshold of $\tau=0.934$, the system maintains a $99.01\%$ Selective Macro-F1 by deferring the most uncertain $33.8\%$ of samples.}
    \label{fig:combined_uncertainty_results}
\end{figure}

    The ResNet-18 baseline, implemented via the \verb:torch-uncertainty: framework \cite{lafage_torch-uncertainty_2025}, demonstrates strong classification performance on the test set. However, as illustrated in the reliability diagrams (Figure \ref{fig:reliability_calibration}), the uncalibrated model was initially significantly overconfident ($\text{ECE}=0.117$). Post-hoc temperature scaling yielded an optimal temperature of $T=2.12$, aligning the model's confidence closely with the identity diagonal and reducing the calibration error to $0.055$. This ensures that the softened softmax outputs are a statistically valid proxy for empirical likelihood, a prerequisite for robust triage.
    
    Beyond point-prediction accuracy, we assess the framework's utility under a rejection policy using metrics that account for taxonomic imbalance (Table \ref{tab:baseline_results}). The Global Macro-F1 score of $90.35\%$ highlights the challenge of rare species; however, by employing selective classification, we significantly enhance the reliability of the automated subset.
    
    \begin{table}[htbp]
        \centering
        \caption{Performance metrics for the calibrated ResNet-18 on the test set.}
        \label{tab:baseline_results}
        \begin{tabular}{lc}
        \hline
        \textbf{Metric} & \textbf{Value} \\ \hline
        Global Accuracy & $95\%$ \\
        Global Macro-F1 & $90.35\%$ \\
        Expected Calibration Error (ECE) & $0.055$ \\
        Area Under the Risk-Coverage Curve (AURC) & $0.0061$ \\
        Selective Macro-F1 (at target operating point) & $99.01\%$ \\
        Coverage (at target operating point) & $66.2\%$ \\
        Optimal Temperature ($T$) & $2.12$ \\ \hline
        \end{tabular}
    \end{table}

    As shown in the F1-coverage analysis (Figure \ref{fig:risk_coverage}), we applied an abstention threshold of $\tau=0.934$ to simulate a high-stakes monitoring deployment. Under this policy, the system automatically processes $66.2\%$ of specimens, achieving a Selective Macro-F1 of $99.01\%$. The remaining $33.8\%$ of ambiguous or rare cases are deferred to a human specialist. This behaviour fulfils the primary objective of the proposed pipeline: providing high-throughput automation for routine specimens while ensuring that cases prone to misclassification are systematically referred for expert review.

   \subsection{Qualitative Analysis of Uncertainty and Triage Utility}

\begin{figure}[htbp]
    \centering
    \includegraphics[width=0.75\textwidth]{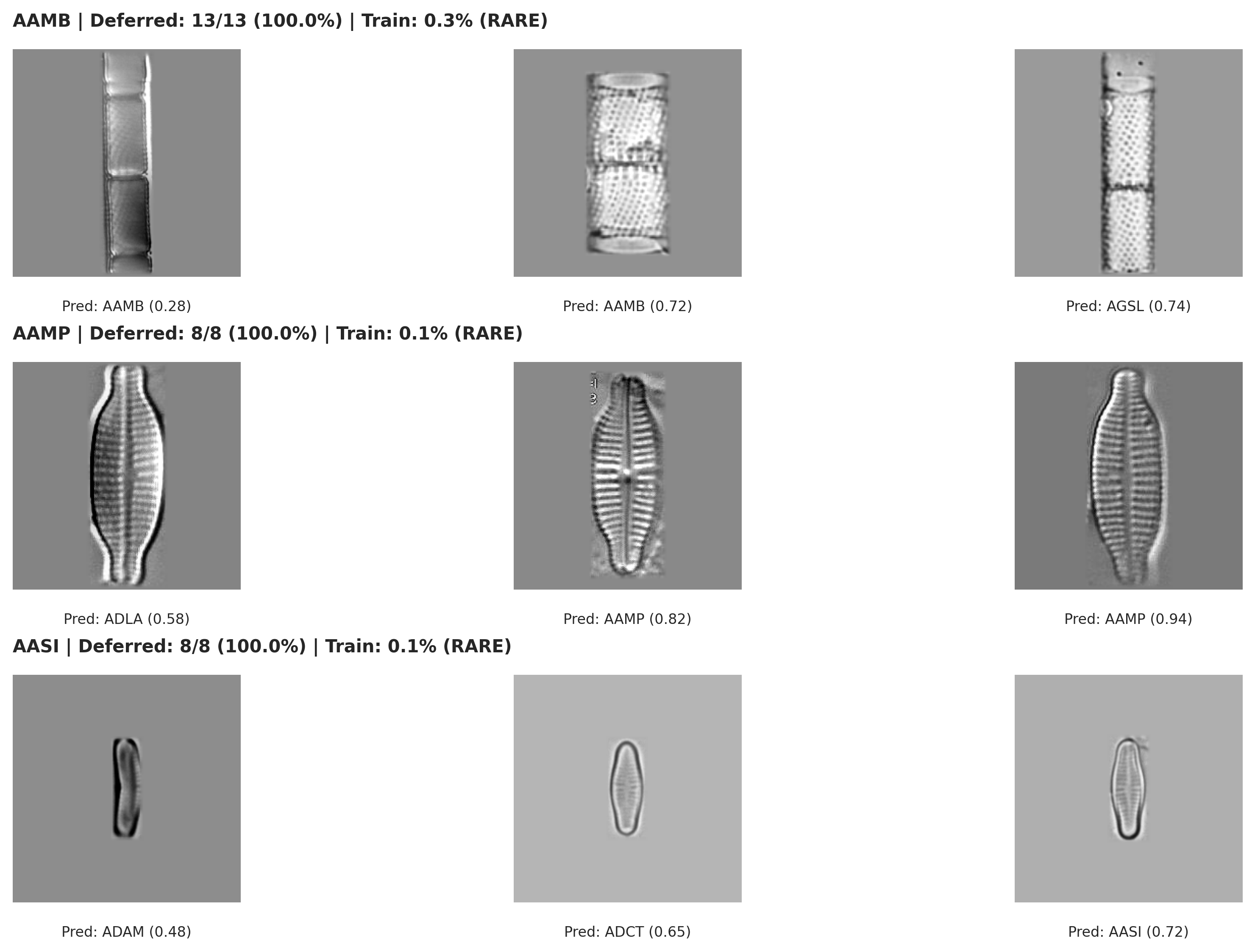}
    \caption{\textbf{Qualitative analysis of deferred specimens across rare taxa.} The figure showcases the lowest-confidence specimens for species frequently targeted for human intervention. Taxa such as AAMB, AAMP, and AASI—all representing less than $0.3\%$ of the training data—exhibit $100\%$ deferral rates. The system successfully flags out-of-distribution morphologies  (e.g., girdle views or small valves), ensuring these high-stakes specimens are prioritized for expert review.}
    \label{fig:deferred_samples}
\end{figure}

To assess the practical utility of the selective classification pipeline for taxonomic workflows, we examined the specimens most frequently deferred by the model at the $\tau=0.934$ operating threshold. As illustrated in Figure \ref{fig:deferred_samples}, the system exhibits a high propensity for rejecting "rare" species—taxa with minimal representation in the training set (e.g., $<0.5\%$).

Specifically, species such as \textit{Aulacoseira ambigua} (AAMB), \textit{Achnanthidium minutissimum} (AAMP), and \textit{Achnanthidium sieminskae} (AASI) were deferred in $100\%$ of test instances. This behavior underscores the model's capacity for epistemic uncertainty detection; rather than forcing a high-confidence prediction for unfamiliar or rare morphologies, the system correctly identifies its own knowledge limits.

From a taxonomist's perspective, this selective mechanism functions as a safeguard for ecological integrity. By isolating ambiguous specimens for human intervention, the pipeline prevents automated misclassification from skewing community indices. This confirms that the calibrated ResNet-18 functions not merely as a rigid classifier, but as an intelligent filter that bridges the gap between high-throughput automated processing and expert domain knowledge.

    \section{Discussion}

The transition from a standard risk-based coverage policy to one governed by the Selective Macro-F1 score represents a fundamental shift toward prioritizing taxonomic integrity over simple data throughput. While global accuracy often masks a model's failure to identify rare bioindicators, the Macro-F1 metric ensures that each taxon, regardless of its frequency in the dataset, is afforded equal weight. Consequently, our calibrated ResNet-18 achieves a Selective Macro-F1 of \textbf{99.01\%}. This level of reliability directly satisfies the rigorous, high-stakes requirements of ecological monitoring.

A primary observation in our qualitative analysis is that the system’s predictive uncertainty aligns closely with genuine taxonomic difficulty. As illustrated by the most frequently deferred specimens (Figure \ref{fig:deferred_samples}), morphological ambiguity—particularly visible in the final row—represents a significant hurdle. Even for seasoned experts, small specimens or those presented in girdle view offer fewer diagnostic features, making confident identification challenging. It is important to clarify that the "rarity" observed here pertains specifically to representation within the current, evolving DIATLAS dataset rather than necessarily reflecting natural abundance within local biodiversity. However, this dataset-driven sparsity directly influences the model's epistemic uncertainty; with fewer examples from which to learn, the model correctly adopts a more conservative, risk-averse posture.

The restrictive nature of optimizing for the Macro-F1 score is inherently more conservative than utilizing a standard accuracy threshold. Because a single error in a rare class catastrophically degrades the macro-average, the resulting rejection threshold ($\tau = 0.934$) is naturally pushed higher. In the context of bioindication, this conservative behavior is highly desirable: the system prefers to defer an ambiguous specimen to a specialist rather than risk a misclassification that could skew sensitive ecological indices. Furthermore, analyzing the subtleties among misclassified cases reveals that many rejected specimens exhibit high visual similarity to overlapping taxa. This suggests that the model successfully identifies regions of the feature space where class boundaries are blurred—a phenomenon we intend to explore further by presenting side-by-side comparisons of misclassified pairs to explicitly showcase these morphological overlaps.

Ultimately, high-quality biodiversity assessment requires moving beyond global risk optimization toward class-conditional logic. While our temperature scaling has effectively reduced the Expected Calibration Error (ECE) to 0.055—providing a reliable confidence signal for triage—the current deterministic approach still conflates data noise (aleatoric uncertainty) with model ignorance (epistemic uncertainty). Future iterations of this framework will integrate explicit second-order uncertainty estimators to better disentangle these sources of error. By adapting rejection logic to both the data density and the morphological difficulty of specific taxa, we can maximize throughput without compromising the integrity of the ecological signal. Over time, incorporating active learning pipelines will further reduce expert workload as the system matures.
    \section{Conclusion}

This work establishes a baseline for uncertainty-aware diatom identification using a ResNet-18 architecture trained on a taxonomically diverse dataset of 759 species, characterised by extreme class imbalance. While the baseline model achieved a high global accuracy, its raw softmax outputs were significantly overconfident ($\text{ECE} = 0.117$). Post-hoc temperature scaling effectively mitigated this miscalibration, aligning predictive confidence with empirical likelihood and reducing the calibration error to $0.055$.\\

The practical utility of these calibrated confidences was demonstrated through a selective classification framework governed by the Macro-F1 score. By prioritising taxonomic integrity over simple point-accuracy, we established a high-reliability regime: at an operating threshold of $\tau = 0.934$, the system achieves a Selective Macro-F1 of 99.01\%. Under this policy, the framework automatically processes 66.2\% of specimens while systematically deferring the most ambiguous and rare cases—often those representing less than $0.5\%$ of the training data—for expert review. This balance underscores the feasibility of using automated triage to reduce specialist workload by two-thirds without compromising the ecological signal required for bioindication.\\

However, several challenges remain for operational deployment. Standard softened softmax outputs, while well-calibrated, do not fully disentangle epistemic model ignorance from aleatoric data noise. Qualitative analysis further revealed that morphological similarities and small specimen sizes continue to pose challenges for both the model and human experts. Future research will focus on benchmarking second-order uncertainty estimators, such as Monte-Carlo Dropout and Conformal Prediction, to provide more nuanced rejection criteria with formal coverage guarantees. By integrating these methods into continuous, human-in-the-loop learning cycles, we move toward a resilient monitoring framework capable of adapting to the shifting taxonomic distributions inherent in aquatic ecosystems.
    \subsubsection*{Acknowledgements}

This project  work was funded  the ANR BIOINDIC-IA ANR-24-CE04-0345.

    \printbibliography
\end{document}